\documentclass{article}

\usepackage{arxiv}

\usepackage[utf8]{inputenc}
\usepackage[T1]{fontenc}
\usepackage{hyperref}
\usepackage{url}
\usepackage{booktabs}
\usepackage{amsfonts}
\usepackage{nicefrac}
\usepackage{microtype}
\usepackage{graphicx}
\usepackage{amsmath}
\usepackage{amssymb}
\usepackage{float}
\usepackage{multirow}

\graphicspath{{./fig/}{./images/}{./JBHI_LXL/}}

\AtBeginDocument{\sloppy}
\title{QDSP: An Interpretable Structured Learning Framework for Predicting Death or Cerebral Palsy in Very Low Birth Weight Infants}

\author{
 Ling Wang \\
  College of Computer Science\\
  Sichuan Normal University\\
  Chengdu, Sichuan, China \\
  \texttt{lingwang@sicnu.edu.cn} \\
   \And
 Xiaolong Li \\
  College of Computer Science\\
  Sichuan Normal University\\
  Chengdu, Sichuan, China \\
  \texttt{20231393034@stu.sicnu.edu.cn} \\
  \And
 Hui Zhou \\
  West China Second University Hospital\\
  Sichuan University\\
  Chengdu, Sichuan, China \\
  \texttt{1103570489@qq.com} \\
  \And
 Jing Shi \\
  West China Second University Hospital\\
  Sichuan University\\
  Chengdu, Sichuan, China \\
  \texttt{shijing@scu.edu.cn} \\
  \And
 Fuhao Zhang \\
  College of Computer Science\\
  Sichuan Normal University\\
  Chengdu, Sichuan, China \\
  \texttt{fuahozhang@stu.sicnu.edu.cn} \\
  \And
 Dapeng Chen $^{*}$ \\
  West China Second University Hospital\\
  Sichuan University\\
  Chengdu, Sichuan, China \\
  \texttt{cdp415@163.com} \\
  \And
 Nan Mu $^{*}$ \\
  College of Computer Science\\
  Sichuan Normal University\\
  Chengdu, Sichuan, China \\
  \texttt{nanmu@sicnu.edu.cn} \\
}
\begin{document}
\maketitle

\begin{abstract}
Very low birth weight infants (VLBWI) are at high risk of mortality and severe neurodevelopmental impairment, including cerebral palsy, yet reliable discharge-time prognostic stratification remains challenging in high-dimensional and data-limited clinical settings. To address this problem, we propose QDSP, an interpretable structured learning framework that integrates Quota-guided Subspace Sampling (QSS) and Differentiable-decision-guided Structure Perception (DSP). The QSS module constructs stability-aware and low-redundancy feature subspaces through bootstrap-based feature consistency estimation, whereas the DSP module employs differentiable soft oblique decision structures to model nonlinear clinical interactions while preserving traceable decision evidence. The proposed framework was evaluated on a real-world VLBWI cohort comprising 51 infants and further validated on three public medical tabular datasets. On the primary cohort, QDSP achieved an accuracy of 0.9200 and an AUC of 0.9714, outperforming representative machine learning and deep tabular learning baselines, including XGBoost, TabNet, and TabPFN. Across external datasets, QDSP maintained competitive discrimination and calibration under varying sample sizes and clinical distributions. In addition, SHAP-based analyses and differentiable decision-path tracing identified clinically relevant predictors, including cystic periventricular leukomalacia (cPVL) and birth weight, consistent with established neonatal pathophysiological evidence. These results suggest that QDSP provides an interpretable and robust framework for discharge-time risk stratification in VLBWI and may support early individualized clinical decision-making in neonatal intensive care settings.
\end{abstract}

\keywords{Very low birth weight infants \and Cerebral palsy \and Interpretable machine learning \and Differentiable decision trees \and Clinical risk stratification \and Small-sample learning}

\section{Introduction}

Very low birth weight infants (VLBWI) are highly vulnerable to mortality and long-term neurodevelopmental impairment because of physiological immaturity and fragile immune function \cite{1}. Early prognostic stratification at hospital discharge is therefore critical for timely intervention and individualized follow-up management aimed at improving long-term clinical outcomes \cite{2}. Recent systematic evidence further indicates that clinically deployable discharge-time prediction tools for long-term neurodevelopmental prognosis in VLBWI remain limited \cite{3}. In practice, neonates who subsequently die during follow-up may initially present discharge characteristics similar to those of infants who survive but later develop severe neurological sequelae such as cerebral palsy. This substantial clinical overlap complicates early risk differentiation and limits the effectiveness of conventional score-based prognostic approaches.

Machine learning has shown increasing potential for neonatal outcome prediction; however, several methodological challenges remain unresolved in this domain \cite{4}. Previous studies have demonstrated that predictive performance in neonatal cohorts is highly sensitive to data quality, cohort heterogeneity, and limited sample size \cite{5}. In high-dimensional clinical tabular settings, unstable feature selection and spurious correlations may substantially impair generalization ability \cite{6}. Meanwhile, highly complex deep learning architectures often sacrifice interpretability, thereby limiting their applicability in safety-critical clinical decision-making scenarios \cite{7}. These limitations highlight the need for modeling frameworks that simultaneously support robust prediction, stable representation learning, and transparent clinical interpretation.

To address these challenges, we propose the Quota-Differentiable Structural Perception (QDSP) framework, which integrates Quota-guided Subspace Sampling (QSS) with Differentiable-decision-guided Structure Perception (DSP). The proposed framework is motivated by recent evidence suggesting that effective inductive bias and structured representation learning are particularly important for deep learning on tabular data, especially under small-sample conditions \cite{7}. Specifically, QSS performs stability-aware feature subspace construction to reduce redundancy and improve representation robustness, whereas DSP employs differentiable soft oblique decision structures to capture nonlinear clinical interactions while preserving explicit decision-path evidence under end-to-end optimization.

The main contributions of this study are summarized as follows:
\begin{enumerate}
\item We propose QDSP, an interpretable structured learning framework for high-dimensional and small-sample clinical tabular prediction that jointly considers predictive discrimination, representation stability, and clinical interpretability.
\item We develop two complementary modules: QSS for stability-aware low-redundancy feature subspace construction and DSP for differentiable structured decision modeling with explicit decision-path tracing capability.
\item We evaluate QDSP on a real-world VLBWI cohort and multiple public medical tabular datasets, demonstrating competitive predictive performance, robustness under limited-data settings, and clinically consistent interpretability compared with representative machine learning and deep tabular learning baselines.
\end{enumerate}

\section{Related Work}

Existing studies on prognostic prediction in VLBWI can generally be categorized into three groups: clinical scoring systems, traditional machine learning methods, and deep learning approaches. These studies have progressively improved neonatal outcome prediction; however, important limitations remain regarding robustness, scalability, and interpretability in small-sample clinical settings.

\subsection{Clinical Scoring Systems}

Early prognostic studies primarily relied on handcrafted clinical scoring systems for mortality and neurodevelopmental risk assessment. Gera and Ramji identified gestational age, birth weight, and mechanical ventilation as important predictors of early mortality in very low birth weight neonates \cite{8}. B\"uhrer et al. further compared CRIB, CRIB-II, gestational age, and birth weight for mortality risk evaluation in VLBWI populations \cite{9}. More recently, a systematic review on neurodevelopmental prediction in premature infants emphasized the continuing clinical need for reliable discharge-time prognostic stratification strategies \cite{3}.

Although these scoring systems provide clinically interpretable risk assessment, most remain based on predefined linear formulations and limited variable interactions. Consequently, their ability to characterize heterogeneous clinical trajectories and complex nonlinear relationships in neonatal intensive care unit (NICU) populations remains constrained.

\subsection{Traditional Machine Learning}

Traditional machine learning methods have increasingly been applied to neonatal outcome prediction because of their stronger nonlinear modeling capability. Han \textit{et al.} utilized machine learning techniques to predict postnatal growth failure in VLBWI and demonstrated the potential utility of data-driven prediction in neonatal care \cite{5}. Shu \textit{et al.} proposed an early prediction framework for mortality and major morbidities in preterm VLBWI neonates \cite{10}, while Lee \textit{et al.} employed a random-forest-based model for mortality risk prediction in premature infants \cite{11}. Bowe \textit{et al.} further investigated machine learning approaches for predicting two-year cognitive outcomes in very preterm infants \cite{12}.

Despite improved predictive discrimination, these methods often exhibit reduced robustness in high-dimensional and limited-sample clinical datasets. In particular, unstable feature selection and sensitivity to sampling variation may adversely affect generalization performance and model reliability.

\subsection{Deep Learning}

Recent advances in deep learning have further improved representation learning capability for neonatal prognosis. He et al. proposed a multi-task and multi-stage transfer learning framework for early neurodevelopmental prediction in very preterm infants \cite{13}. Ihlen \textit{et al.} applied machine learning to infant spontaneous movement analysis for early cerebral palsy prediction in a multi-site cohort \cite{14}, and Groos \textit{et al.} developed a deep learning framework for cerebral palsy risk estimation from spontaneous infant movements \cite{15}. These studies demonstrate the potential of deep architectures to capture complex developmental patterns from biomedical data.

However, most existing deep learning approaches focus primarily on unstructured modalities such as imaging or movement signals and provide limited transparency for heterogeneous discharge-time clinical tabular data. Moreover, recent studies on tabular deep learning suggest that effective inductive bias and structured representation learning are essential for improving stability and generalization under small-sample conditions \cite{7}. These observations motivate the incorporation of structured decision priors into our proposed DSP module.

Overall, existing studies demonstrate substantial progress from conventional scoring systems to machine learning and deep learning approaches. Nevertheless, achieving robust prediction, stable representation learning, and clinically interpretable decision evidence simultaneously remains a significant challenge for high-dimensional and small-sample neonatal clinical prediction tasks.

\section{Method}

\subsection{Overall Structure}
\label{sec:overall_structure}

\begin{figure*}[!t]
\centering
\includegraphics[width=0.95\textwidth]{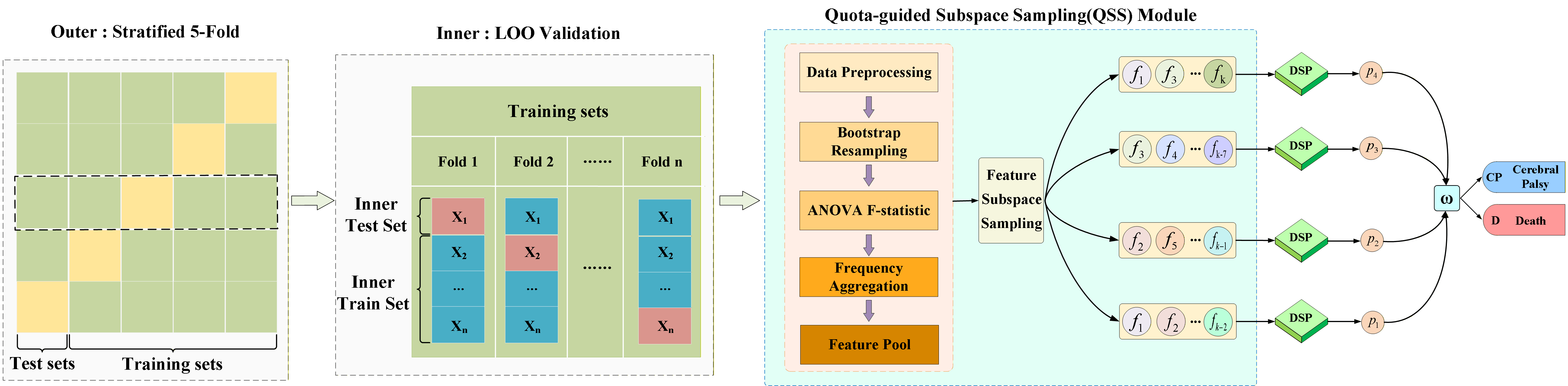}
\caption{Overall architecture of the proposed QDSP framework under the nested validation protocol.}
\label{fig1}
\end{figure*}

Figure~\ref{fig1} presents the overall architecture and training pipeline of the proposed QDSP framework. To obtain reliable performance estimation under limited sample conditions, the framework adopts a nested cross-validation strategy. Specifically, stratified five-fold cross-validation is employed in the outer loop for generalization evaluation, whereas leave-one-out cross-validation (LOOCV) is performed within the inner loop for model training and hyperparameter optimization. This nested design minimizes information leakage while maximizing the utilization of scarce clinical samples.

For each training iteration, the original feature space is first processed by the Quota-guided Subspace Sampling (QSS) module to construct multiple stability-aware and low-redundancy feature subspaces. Each subspace is subsequently forwarded to an independent Differentiable-decision-guided Structure Perception (DSP) submodel, which outputs the corresponding predictive probability. The final prediction is obtained through ensemble aggregation across all DSP submodels, yielding the estimated risk probability for discriminating between the two mutually exclusive adverse outcomes, cerebral palsy and mortality.

\subsection{Quota-guided Subspace Sampling (QSS) Module}

The QSS module is designed to improve feature robustness and subspace diversity under high-dimensional and small-sample clinical conditions. The module consists of two stages: stability-aware feature selection and weighted feature subspace construction.

\subsubsection{Robust Feature Selection and Redundancy Filtering}

The preprocessing pipeline first applies median imputation for missing values, Z-score normalization for continuous variables, and removal of near-zero-variance features to reduce noise and improve modeling stability. Within each training fold, feature stability under sampling perturbation is evaluated using bootstrap resampling with $B = 120$ iterations and a sampling ratio of $\alpha = 0.7$.

Following the stable feature selection strategy proposed by Alelyani \cite{16}, repeated bootstrap sampling and feature ranking are utilized to improve the robustness of feature importance estimation. For each bootstrap subset $D_b$, ANOVA F-statistics are computed for all candidate features, and the top-$k$ features are retained. Let $c_j$ denote the number of times feature $j$ is selected across all bootstrap iterations. The stability frequency of feature $j$ is defined as:
\begin{equation}
    freq_j = \frac{c_j}{B}, \quad j = 1, \dots, p.
\end{equation}

The resulting stability frequency reflects the consistency of feature relevance under small-sample perturbations. Features are subsequently ranked according to $freq_j$, and the top $K = 2.5k$ features are retained to form the initial candidate set $C$.

To further reduce redundancy, a pairwise correlation matrix $R$ is computed within $C$. For any feature pair satisfying $R_{ij} > \tau = 0.9$, only the feature with the higher stability frequency is preserved. This procedure produces a refined low-redundancy feature pool $C^*$ for subsequent subspace construction.

\subsubsection{Weighted Subspace Sampling and Ensemble Configuration}

To generate diverse yet stable feature representations, weighted sampling without replacement is performed on the refined feature pool $C^*$. The sampling probability of feature $j$ is defined according to its stability frequency:
\begin{equation}
    \pi_j = \frac{freq_j}{\sum_{j \in C^*} freq_j}.
\end{equation}

Based on the sampled features, a stability-aware subspace score is further defined as:
\begin{equation}
    score(S_m) = \frac{1}{|S_m|} \sum_{j \in S_m} freq_j.
\end{equation}

Using this strategy, $M = 4$ feature subspaces $S_m \subset C^*$ are generated, where each subspace contains $k$ features under a global feature-usage constraint designed to preserve inter-subspace diversity. Each subspace corresponds to an independent base learner $h_m$, which outputs the predictive probability $p_m(x)$ for input sample $x$.

In principle, the final ensemble prediction can be formulated as:
\begin{equation}
    p(x) = \sum_{m=1}^{M} \alpha_m p_m(x),
    \quad
    \text{s.t.}
    \quad
    \sum_{m=1}^{M} \alpha_m = 1,
    \;
    \alpha_m \geq 0,
\end{equation}
where the ensemble weight is determined by the stability score of the corresponding feature subspace:
\begin{equation}
    \alpha_m =
    \frac{score(S_m)}
    {\sum_{r=1}^{M} score(S_r)}.
\end{equation}

However, preliminary experiments showed that stability-aware weighting did not consistently improve predictive performance under the current small-sample setting. Therefore, an equally weighted ensemble strategy was ultimately adopted, with $\alpha_m = \frac{1}{M}$. This design simplifies the ensemble structure while reducing variance sensitivity and potential overfitting.

\subsection{Differentiable-decision-guided Structure Perception (DSP) Module}

\begin{figure}[!t]
\centering
\includegraphics[width=\linewidth]{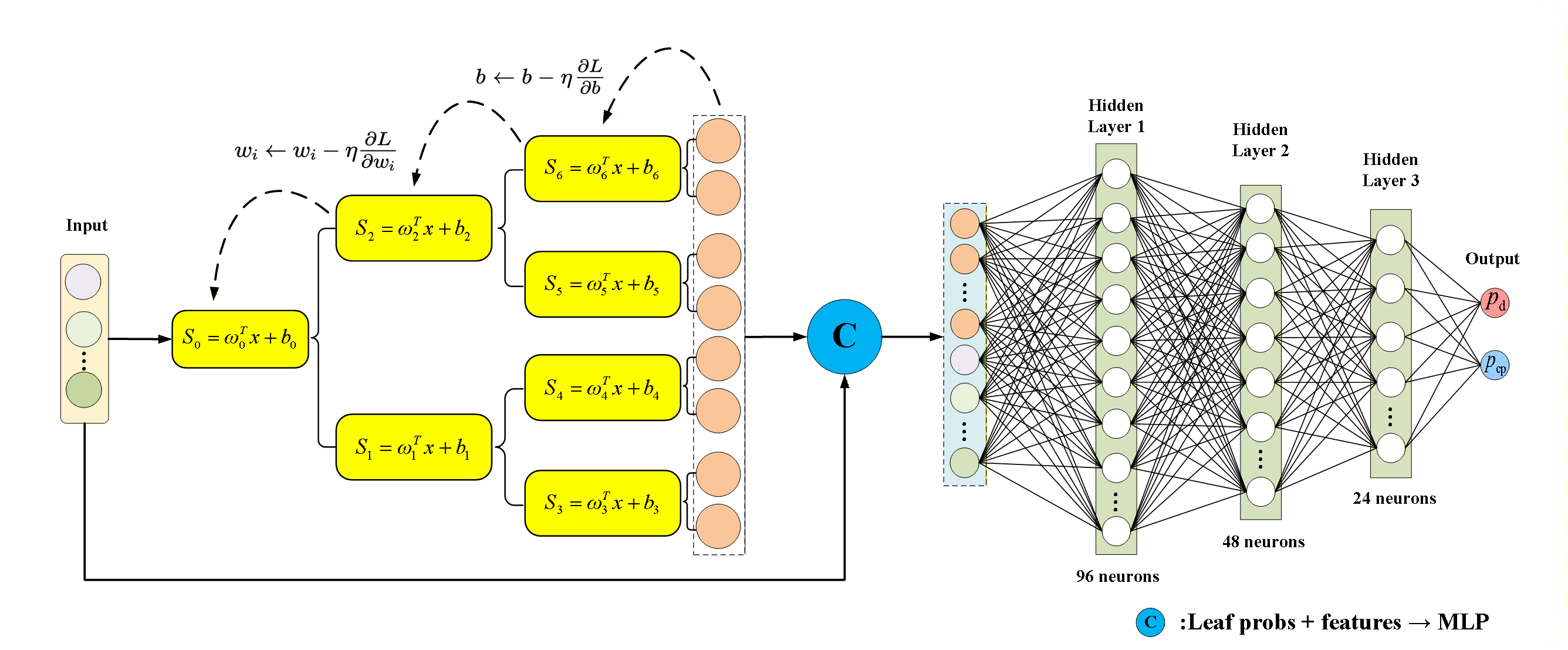}
\caption{Architecture of the proposed DSP module.}
\label{fig2}
\end{figure}

Following the stability-aware feature subspace construction performed by QSS, each selected feature subspace is forwarded to the Differentiable-decision-guided Structure Perception (DSP) module for structured representation learning. As illustrated in Fig.~\ref{fig2}, DSP introduces explicit structural inductive bias into clinical tabular modeling through differentiable oblique decision structures. The module jointly captures hierarchical decision semantics and nonlinear feature interactions while preserving interpretable probabilistic routing evidence.

For each feature subspace $S_m$, a differentiable oblique decision tree with depth $d=3$ is constructed, resulting in $2^d-1$ internal decision nodes and $2^d$ leaf nodes. Given an input feature vector $\mathbf{x} \in \mathbb{R}^p$, the split score at the $j$-th internal node is computed as:
\begin{equation}
    s_j = \boldsymbol{\omega}_j^T \mathbf{x} + b_j,
\end{equation}
where $\boldsymbol{\omega}_j \in \mathbb{R}^{p}$ denotes the learnable oblique projection vector and $b_j$ is the learnable bias term.

Unlike conventional axis-aligned decision rules, the oblique split formulation enables the model to capture coupled interactions among multiple clinical variables, thereby improving representational flexibility for heterogeneous clinical patterns. To maintain end-to-end differentiability, a soft probabilistic routing mechanism is introduced using a sigmoid activation with temperature parameter $\tau = 1.0$:
\begin{equation}
    p_{\text{left}}^{(j)}
    =
    \sigma\left(\frac{s_j}{\tau}\right),
    \quad
    p_{\text{right}}^{(j)}
    =
    1 - p_{\text{left}}^{(j)}.
\end{equation}

Instead of assigning each sample to a single branch deterministically, DSP propagates samples probabilistically through both branches. This soft-routing mechanism preserves gradient flow during optimization while allowing the model to represent uncertainty in clinical decision boundaries. The probability that a sample reaches leaf node $l$ is computed as:
\begin{equation}
    P_l(\mathbf{x})
    =
    \prod_{j \in \text{path}(l)}
    q_j(\mathbf{x}),
\end{equation}
where $q_j(\mathbf{x})$ denotes the routing probability associated with the branch leading toward leaf node $l$.

Aggregating all leaf probabilities yields the structured leaf representation vector:
\begin{equation}
    \mathbf{P}_{\text{leaf}}
    \in
    \mathbb{R}^{2^d},
\end{equation}
which encodes hierarchical probabilistic decision semantics for the input sample.

Although the structured leaf representation provides informative decision priors, relying solely on hierarchical compression may reduce fine-grained feature fidelity. Therefore, DSP further integrates the leaf representation with the original feature vector:
\begin{equation}
    \mathbf{Z}
    =
    [\mathbf{P}_{\text{leaf}}; \mathbf{x}]
    \in
    \mathbb{R}^{2^d+p},
\end{equation}
where $[\cdot ; \cdot]$ denotes vector concatenation.

The fused representation $\mathbf{Z}$ is subsequently forwarded to a multilayer perceptron (MLP) with ReLU activation and Dropout regularization to model higher-order nonlinear relationships. The final risk probability is computed as:
\begin{equation}
    \hat{y}
    =
    \sigma(f_{\text{MLP}}(\mathbf{Z}))
    \in
    (0,1).
\end{equation}

Through this design, DSP combines structured probabilistic decision modeling with nonlinear representation learning, enabling the framework to maintain predictive flexibility while preserving interpretable decision-path evidence for clinical risk analysis.

\subsection{Loss Function}

To address outcome imbalance during optimization, the model is trained using a weighted binary cross-entropy loss. By assigning a larger penalty to minority-class samples, the loss function improves sensitivity to underrepresented outcomes while maintaining stable optimization behavior.

Given training samples $\{x_i\}_{i=1}^{N}$, let $y_i \in \{0,1\}$ denote the ground-truth label, where $y_i=1$ corresponds to cerebral palsy and $y_i=0$ corresponds to death. Let $\hat{y}_i \in (0,1)$ denote the predicted probability of cerebral palsy. The weighted binary cross-entropy loss is defined as:
\begin{equation}
\mathcal{L}
=
-\frac{1}{N}
\sum_{i=1}^{N}
\left[
\omega_{\text{pos}}
\cdot
 y_i
\cdot
\log(\hat{y}_i)
+
(1-y_i)
\cdot
\log(1-\hat{y}_i)
\right],
\end{equation}
where $\omega_{\text{pos}}$ represents the weighting factor for the minority outcome class. The weight is determined according to the class distribution:
\begin{equation}
\omega_{\text{pos}}
=
\frac{N_{\text{neg}}}{N_{\text{pos}}},
\end{equation}
where $N_{\text{neg}}$ and $N_{\text{pos}}$ denote the numbers of negative and positive samples in the training set, respectively.

Compared with standard binary cross-entropy, this formulation imposes a larger optimization penalty on minority-outcome misclassification, thereby improving learning balance under imbalanced clinical data distributions.

\subsection{Design of Interpretability Analysis Methods}

\begin{figure}[!t]
\centering
\includegraphics[width=\linewidth]{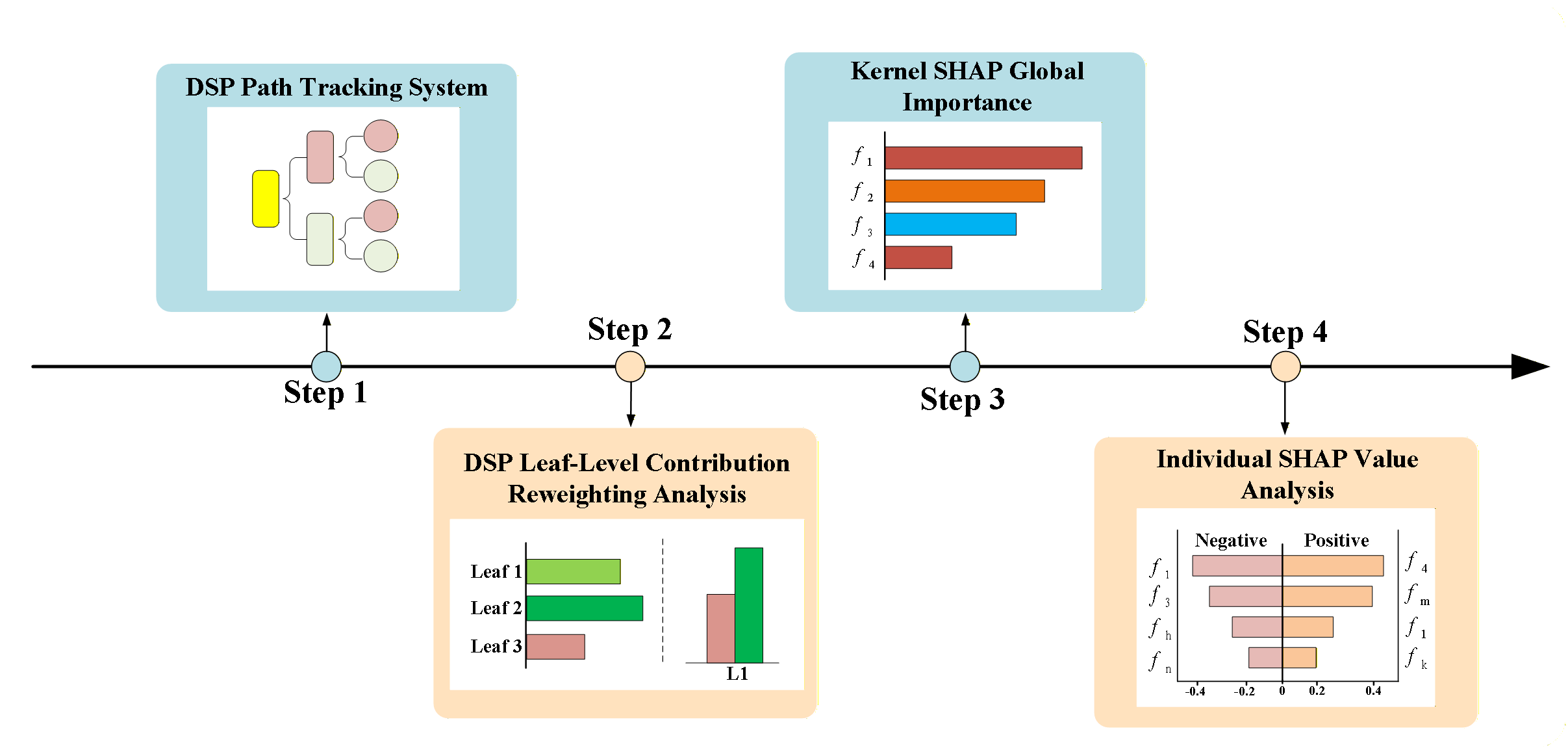}
\caption{Multi-level interpretability framework of the proposed model.}
\label{fig3}
\end{figure}

As illustrated in Fig.~\ref{fig3}, a four-stage interpretability framework was developed to analyze the decision behavior of the proposed model across multiple representation levels. The framework links probabilistic routing mechanisms within DSP to both global and patient-specific feature attribution patterns, thereby providing interpretable evidence from internal decision semantics to final prediction outcomes.

The first stage performs differentiable decision-path tracing at the sample level to reconstruct the probabilistic routing process within DSP. For an input vector $\mathbf{x}$ at the $j$-th internal node, the routing probabilities are defined as:
\begin{equation}
p_{\text{left}}
=
\sigma\left(\frac{s_j}{\tau}\right),
\quad
p_{\text{right}}
=
1-p_{\text{left}},
\end{equation}
where $\tau$ denotes the temperature parameter controlling routing smoothness.

Unlike deterministic tree traversal, the soft-routing mechanism generates probabilistic decision trajectories that preserve uncertainty information during inference. For visualization, routing tendencies are represented using directional color coding, where green indicates left-dominant routing and red indicates right-dominant routing. To improve interpretability at the node level, the five most influential features are identified according to the absolute magnitudes of the internal projection weights.

The second stage evaluates leaf-level contribution behavior and routing uncertainty. To quantify the predictive sensitivity of each leaf node, a perturbation-based analysis is performed using a probability perturbation $\Delta p = 0.1$:
\begin{equation}
\nabla_k
=
\frac{
\text{MLP}(p_k+\Delta p,\mathbf{x})
-
\text{MLP}(p_k,\mathbf{x})
}
{\Delta p},
\end{equation}
where $\nabla_k$ represents the sensitivity of the predicted outcome to perturbation at leaf node $k$, and $p_k$ denotes the original routing probability of that leaf.

To further characterize routing uncertainty, decision dispersion across leaf nodes is quantified using Shannon entropy and Gini impurity:
\begin{equation}
H
=
-\sum_{k=1}^{L}
p_k \log_2(p_k),
\quad
G
=
1-\sum_{k=1}^{L}p_k^2,
\end{equation}
where $L$ denotes the total number of leaf nodes and $p_k$ represents the probability mass assigned to leaf $k$.

An amplification ratio is further defined as:
\begin{equation}
\alpha_k
=
\frac{R_k}{p_k},
\end{equation}
where $R_k$ denotes the perturbed contribution score of leaf $k$. Large amplification ratios indicate that certain leaves contribute disproportionately to the final prediction relative to their routing probabilities, reflecting the collaborative multi-leaf decision behavior of DSP.

The third and fourth stages employ SHAP to analyze feature attribution at both the population and individual levels. For global interpretation, the average absolute SHAP value across all evaluated samples is computed as:
\begin{equation}
\Phi_j
=
\frac{1}{n}
\sum_{i=1}^{n}
|\phi_j^{(i)}|,
\end{equation}
where $\Phi_j$ denotes the global importance score of feature $j$, $n$ is the number of evaluated samples, and $\phi_j^{(i)}$ is the SHAP value of feature $j$ for sample $i$.

At the patient level, SHAP analysis is further applied to representative high-confidence prediction cases, including true-positive and true-negative samples. The most influential positive and negative contributing features are visualized to identify the dominant clinical factors associated with specific prediction outcomes. Together, these analyses establish a unified interpretability hierarchy spanning internal routing dynamics, leaf-level contribution behavior, population-level dependencies, and patient-specific decision evidence.

\section{Experiments}

\subsection{Dataset}

\textbf{Primary Clinical Dataset: VLBWI Neonatal Follow-up Cohort.}
The primary clinical cohort was retrospectively collected from the electronic medical record (EMR) system of the Department of Neonatology, West China Second University Hospital, Sichuan University, spanning the period from 2019 to 2024. Clinical variable extraction and coding were independently conducted by trained neonatology clinicians according to a standardized data-processing protocol.

Guided by prior neonatal prognostic studies and the CHNN Hospital Born Neonate Database, a multi-domain feature system was established, including prenatal characteristics, perinatal indicators, neonatal complications, and treatment-related variables. The inclusion criteria were as follows: 1) very low birth weight infants ($BW < 1500$ g); 2) complete discharge-time EMR data; 3) receipt of standardized neonatal intensive care; and 4) a definitive long-term outcome of either death or cerebral palsy (CP) confirmed during follow-up evaluation.

The final cohort included 51 infants, consisting of 37 mortality cases and 14 CP cases, with a total of 53 clinical variables retained for analysis. To preserve the temporal consistency of prognostic modeling, only variables available up to hospital discharge were included. Consequently, the proposed framework performs long-term outcome prediction strictly based on discharge-time clinical characteristics.

Although the cohort size is limited because of the strict inclusion criteria and the rarity of clinically confirmed long-term outcomes in VLBWI populations, the dataset provides a clinically rigorous setting for evaluating small-sample prognostic modeling under real-world neonatal conditions.

\textbf{Public Benchmarking Datasets: Generalization Evaluation.}
To further evaluate the methodological generalizability of QDSP across heterogeneous biomedical tabular scenarios, three publicly available medical datasets were additionally included for external benchmarking.

The UCI Heart Disease Dataset \cite{17} contains 303 patient records from the Cleveland Clinic database and is designed for heart disease prediction using 13 clinical variables. The Pima Indians Diabetes Dataset \cite{18} includes 768 patient records for diabetes onset prediction based on metabolic and physiological indicators. The Stroke Prediction Dataset\footnote{\url{https://www.kaggle.com/datasets/fedesoriano/stroke-prediction-dataset}} consists of 5,110 samples for stroke risk prediction using demographic, lifestyle, and clinical attributes. For the stroke dataset, majority-class undersampling was applied during training to mitigate severe class imbalance.

These external datasets were used to evaluate the robustness and transferability of QDSP under varying sample sizes, feature distributions, and clinical prediction tasks, rather than for direct neonatal clinical validation.

\subsection{Experimental Configuration}

\textbf{Environment Configuration.}
All experiments were implemented in Python 3.10 using the PyTorch 2.3.1 framework. Model training and evaluation were conducted on a workstation equipped with an Intel Core processor and 16 GB RAM. To ensure reproducibility, the random seed was fixed at 42 for all stochastic operations and model initialization procedures.

\textbf{Training Strategy and Hyperparameters.}
Considering the limited sample size of the primary clinical cohort, model development was performed under the nested cross-validation framework described in Section~\ref{sec:overall_structure} to maximize data utilization while providing reliable performance estimation. Specifically, leave-one-out cross-validation (LOOCV) was adopted in the inner training loop for model optimization and threshold selection. The optimal decision threshold was determined by maximizing balanced accuracy within the training folds and subsequently fixed for evaluation on the held-out outer-loop samples. This strategy preserves strict separation between threshold optimization and final testing. The model was optimized using the AdamW optimizer with an initial learning rate of $1 \times 10^{-3}$ and a maximum training duration of 150 epochs.

\textbf{Evaluation Metrics.}
To comprehensively assess predictive performance and clinical reliability, six evaluation metrics were adopted: Accuracy, Precision, Recall, F1-score, Area Under the Receiver Operating Characteristic Curve (AUC), and Brier score.

Accuracy evaluates overall classification correctness, whereas Precision measures the reliability of positive predictions and Recall quantifies the ability to identify high-risk cases. The F1-score provides a balanced assessment of Precision and Recall, which is particularly informative under imbalanced clinical outcome distributions. AUC was used as the primary metric for evaluating discriminative capability across different classification thresholds. In addition, the Brier score was employed to assess probability calibration quality:
\begin{equation}
\text{Brier}
=
\frac{1}{N}
\sum_{i=1}^{N}
(\hat{y}_i-y_i)^2.
\end{equation}

A lower Brier score indicates better agreement between predicted probabilities and observed clinical outcomes, reflecting improved reliability for clinical risk stratification and decision support.

\subsection{Comparison with State-of-the-Art Models}

To evaluate the effectiveness of the proposed framework, QDSP was compared with several representative machine learning and deep tabular learning methods, including Logistic Regression \cite{19}, Support Vector Machine (SVM) \cite{20}, Decision Tree \cite{21}, Random Forest \cite{22}, LightGBM \cite{23}, XGBoost \cite{24}, TabNet \cite{25}, and TabPFN \cite{26}.

To ensure fair comparison, all datasets were processed using the same preprocessing pipeline and evaluated under identical nested cross-validation protocols. All preprocessing operations, hyperparameter optimization procedures, and threshold selection steps were performed exclusively within the training folds. Test data were not involved in model fitting or threshold optimization. We first evaluated QDSP against all baseline methods on the primary VLBWI cohort. The comparative results are summarized in Table~\ref{tab:model_performance}.

\begin{table}[!t]
\centering
\caption{Performance comparison of different models on the primary VLBWI cohort.}
\label{tab:model_performance}
\small
\setlength{\tabcolsep}{4pt}
\renewcommand{\arraystretch}{1.1}
\resizebox{\textwidth}{!}{%
\begin{tabular}{lcccccc}
\toprule
Model & Accuracy$\uparrow$ & Precision$\uparrow$ & Recall$\uparrow$ & AUC$\uparrow$ & F1-Score$\uparrow$ & Brier Score$\downarrow$ \\
\midrule
Logistic Regression & 0.7855 & 0.8067 & 0.9179 & 0.8000 & 0.8575 & 0.2066 \\
SVM                 & 0.7455 & 0.7432 & \textbf{1.0000} & 0.7345 & 0.8508 & 0.1925 \\
Decision Tree       & 0.7055 & 0.8384 & 0.7821 & 0.7506 & 0.7808 & 0.1586 \\
Random Forest       & 0.8036 & 0.8379 & 0.9143 & 0.8036 & 0.8697 & 0.1876 \\
LightGBM            & \underline{0.8818} & 0.9056 & \underline{0.9429} & 0.9339 & 0.9165 & 0.0999 \\
XGBoost             & 0.7255 & 0.7255 & \textbf{1.0000} & 0.6494 & 0.8403 & 0.2454 \\
TabNet              & 0.6000 & 0.5427 & 0.6571 & 0.8955 & 0.5818 & 0.2417 \\
TabPFN              & 0.8812 & \underline{0.9220} & 0.9250 & \underline{0.9500} & \underline{0.9213} & \underline{0.0772} \\
QDSP (Ours)         & \textbf{0.9200} & \textbf{0.9528} & \underline{0.9429} & \textbf{0.9714} & \textbf{0.9416} & \textbf{0.0658} \\
\bottomrule
\end{tabular}}
\end{table}

As shown in Table~\ref{tab:model_performance}, conventional machine learning models demonstrated limited overall performance on the small-sample VLBWI cohort. Among them, Random Forest achieved the strongest results, with an accuracy of 0.8036 and an F1-score of 0.8697. Boosting-based approaches generally yielded improved discriminative performance, particularly LightGBM, which achieved an AUC of 0.9339 and a Brier score of 0.0999. TabPFN also demonstrated strong performance, obtaining an AUC of 0.9500 and the second-best Brier score of 0.0772. In contrast, TabNet showed unstable behavior under the current limited-data setting despite relatively high AUC performance.

Overall, QDSP achieved the best balance between discrimination and calibration across all evaluated metrics. Specifically, QDSP obtained an accuracy of 0.9200, an AUC of 0.9714, an F1-score of 0.9416, a precision of 0.9528, and a recall of 0.9429. In addition, QDSP achieved the lowest Brier score (0.0658), indicating improved probability calibration consistency relative to the competing approaches.

To further assess methodological generalizability, all methods were additionally evaluated on three external public medical tabular datasets, including Heart Disease, Pima Diabetes, and Stroke, using the same preprocessing and nested cross-validation protocol. For the Stroke dataset, majority-class undersampling was applied during training to mitigate severe outcome imbalance. Detailed quantitative results are provided in \textbf{Appendix Tables~\ref{tab:app_heart_performance}--\ref{tab:app_stroke_performance}}.

Across the external datasets, QDSP maintained competitive generalization performance under varying sample sizes and clinical prediction scenarios. QDSP achieved the highest classification accuracy on all three datasets, together with competitive discriminative performance on Heart Disease (AUC = 0.9036) and Stroke (AUC = 0.8371). In addition, QDSP obtained the best Brier scores on both Heart Disease (0.1086) and Stroke (0.1666), suggesting favorable calibration behavior across heterogeneous clinical distributions. On Pima Diabetes, QDSP maintained competitive discrimination performance (F1-score = 0.6809; AUC = 0.8212), although probability calibration was relatively weaker with a Brier score of 0.1908. Overall, these findings support the transferability of QDSP across diverse medical tabular prediction tasks while indicating that calibration behavior remains partially dataset-dependent.

\subsection{Evaluation of Small-Sample Robustness}

Because medical tabular datasets are often constrained by limited sample availability, performance evaluated under a single full-training setting may overestimate model generalization and mask potential overfitting. To further investigate whether the performance advantage of QDSP reflects genuine adaptability to small-sample conditions, we conducted an additional robustness analysis on the Heart Disease dataset. Specifically, while keeping the test set fixed, the training set size was progressively reduced to 100\%, 80\%, 60\%, 40\%, 30\%, and 20\% of the original training data. Model performance was then tracked across these progressively data-constrained settings to evaluate stability and generalization under limited-sample conditions. This analysis provides additional evidence regarding the reliability and practical applicability of QDSP in small-scale clinical learning scenarios.

\begin{figure}[!t]
\centering
\includegraphics[width=0.95\linewidth]{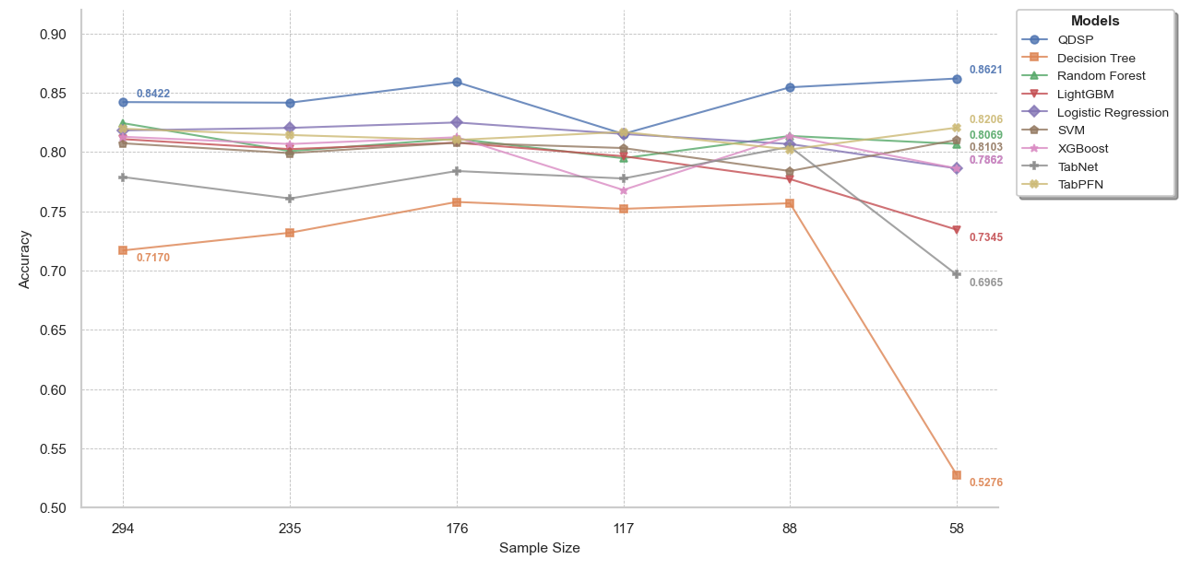}
\caption{Accuracy trends under progressively reduced training sample sizes.}
\label{fig4}
\end{figure}

Figure~\ref{fig4} illustrates the classification accuracy of all compared models as the training set size decreases. Overall, QDSP maintains relatively stable predictive performance across different sampling ratios. Even when only 20\% of the original training data are retained, QDSP still achieves an accuracy of 0.8621, remaining competitive with or outperforming several representative baseline models. This result suggests that the proposed framework preserves stable decision boundaries under severe sample scarcity.

\begin{figure}[!t]
\centering
\includegraphics[width=0.95\linewidth]{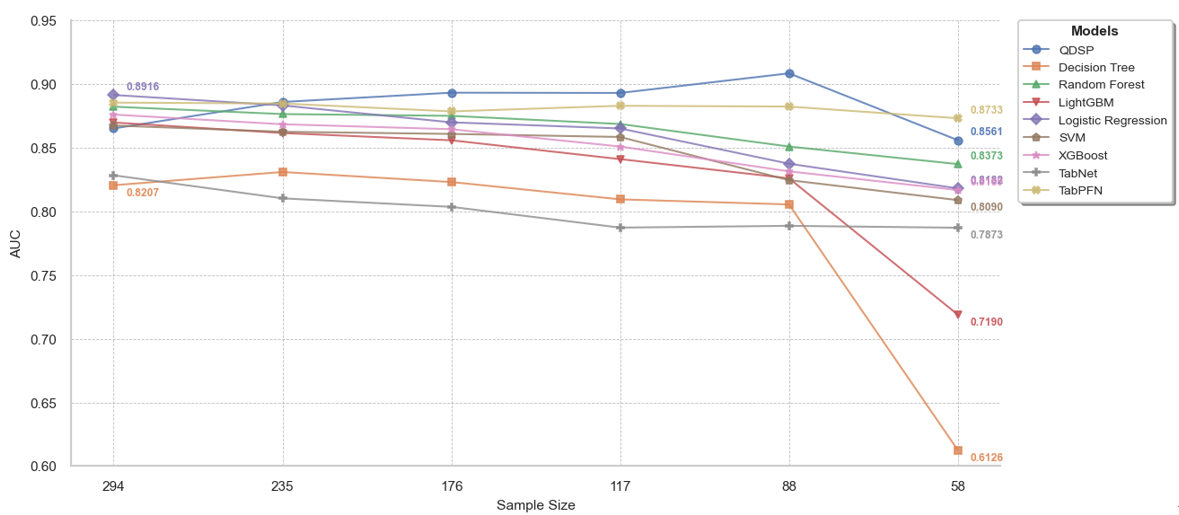}
\caption{AUC trends under progressively reduced training sample sizes.}
\label{fig5}
\end{figure}

Figure~\ref{fig5} presents the corresponding AUC results under progressively reduced training sizes. QDSP consistently maintains competitive discriminative capability across different sample scales and achieves an AUC of 0.9086 at the 30\% training ratio. Compared with several baseline approaches whose discriminative performance degrades substantially under limited-data conditions, QDSP demonstrates stronger robustness and preserved ranking capability in data-scarce scenarios.

Taken together, the results in Figs.~\ref{fig4} and~\ref{fig5} indicate that QDSP maintains both classification accuracy and discriminative performance as the available training data decrease. These findings further support the suitability of the proposed framework for high-dimensional, small-sample clinical tabular prediction tasks, where stable generalization under limited data availability is particularly critical.

\subsection{Ablation Study}

To investigate the individual contribution of each component and their synergistic effects, we conducted an ablation study under the same data split and evaluation protocol as the full model. Table~\ref{tab:ablation_study} reports the performance of different model variants on the primary VLBWI dataset. The compared settings are defined as follows: (1) \textbf{B+MLP}: a baseline model using raw features with a standard MLP classifier, which serves as a reference for performance without explicit feature selection or structured decision modeling. (2) \textbf{B+DSP ($-$LPF)}: a variant incorporating the DSP module while removing the Leaf Probability Fusion (LPF) mechanism, where leaf-node probabilities are not fused with raw features, allowing assessment of DSP's standalone contribution. (3) \textbf{B+DSP ($+$LPF)}: based on the previous variant, LPF is enabled to integrate DSP-derived leaf probabilities with original features, evaluating the effect of probabilistic feature fusion. (4) \textbf{B+QSS+MLP}: a model that introduces the QSS module for stable feature subspace construction while retaining an MLP prediction head, isolating the contribution of feature-level robustness without structured decision modeling. (5) \textbf{B+QSS+DSP (Ours)}: the full QDSP framework, combining QSS and DSP with LPF enabled by default, enabling joint optimization of robust feature selection, structured decision modeling, and probabilistic feature fusion.

\begin{table}[!t]
\centering
\caption{Ablation study results on the primary VLBWI dataset.}
\label{tab:ablation_study}
\small
\setlength{\tabcolsep}{4pt}
\renewcommand{\arraystretch}{1.1}
\resizebox{\textwidth}{!}{%
\begin{tabular}{lcccccc}
\toprule
Model & Accuracy$\uparrow$ & Precision$\uparrow$ & Recall$\uparrow$ & AUC$\uparrow$ & F1-Score$\uparrow$ & Brier Score$\downarrow$ \\
\midrule
B+MLP              & 0.8636 & 0.8769 & \textbf{0.9464} & 0.9380 & \underline{0.9080} & 0.1162 \\
B+DSP($-$LPF)      & 0.8318 & 0.8996 & 0.8857 & 0.9476 & 0.8655 & 0.0939 \\
B+DSP($+$LPF)      & 0.8800 & \underline{0.9750} & 0.8679 & \textbf{0.9714} & 0.9072 & \textbf{0.0396} \\
B+QSS+MLP          & \underline{0.9000} & \textbf{0.9778} & 0.9143 & 0.9075 & \underline{0.9337} & \underline{0.0888} \\
B+QSS+DSP (Ours)   & \textbf{0.9200} & 0.9528 & \underline{0.9429} & \textbf{0.9714} & \textbf{0.9416} & 0.0658 \\
\bottomrule
\end{tabular}}
\end{table}

As seen in Table~\ref{tab:ablation_study}, the results indicate that each component contributes complementary benefits to the overall performance. The DSP module improves discriminative modeling, particularly in terms of AUC, and provides a structured decision mechanism that enhances predictive consistency. The introduction of LPF further improves probability calibration and overall classification performance by effectively fusing leaf-level decision information with raw feature representations. The QSS module contributes to more stable and informative feature subspaces, leading to consistent improvements in both accuracy and F1-score. Overall, the full QDSP model achieves the most balanced performance across all metrics, demonstrating that combining robust feature selection with differentiable structured decision modeling yields superior generalization in small-sample clinical settings. A slight trade-off in calibration is observed compared with the DSP+LPF variant, suggesting that the integrated model produces slightly sharper probability estimates under highly constrained sample conditions.

\subsection{Analysis of Interpretability Results}

From a structural and fine-grained perspective, we first analyze a representative correctly classified case with high confidence, where the ground-truth label is 1 and the predicted probability is 0.9997. The corresponding decision process within the DSP module is visualized in \textbf{Appendix Fig.~\ref{fig:app_decision_trace}}. Unlike conventional hard-decision trees, DSP employs a differentiable soft-splitting mechanism, where each internal node routes the input sample to both left and right branches according to probabilistic weights. Therefore, the highlighted routing path in the visualization represents the dominant splitting tendency rather than a deterministic decision path, improving interpretability while preserving full gradient-based optimization.

As illustrated in \textbf{Appendix Fig.~\ref{fig:app_decision_trace}}, the sample traverses a depth-3 DSP structure through three successive soft splits and ultimately accumulates the highest probability mass at Leaf~2 (0.170). At each internal node, the model simultaneously records split scores, branch probabilities, and feature-wise contributions, enabling explicit tracing of local decision evidence. Across the first and second layers, the dominant features exhibit consistent directional effects, jointly pushing the sample toward the left subtree. However, at the third layer, partial feature contributions shift the routing preference, resulting in redistributed probability mass across multiple leaves. Notably, although Leaf~2 holds the highest probability, other leaves (e.g., Leaf~5) also receive non-negligible probability, reflecting the intrinsic multi-path nature of DSP. This cooperative leaf activation mechanism enables the model to retain complementary decision evidence across different branches rather than relying on a single deterministic trajectory.

To further investigate the multi-leaf decision behavior, we analyze \textbf{Appendix Fig.~\ref{fig:app_multi_leaf}} by comparing leaf-wise probabilities with perturbation-based contribution strengths. The results indicate that final predictions are not solely determined by raw routing probabilities. In particular, some leaves with relatively lower probabilities exhibit higher sensitivity under perturbation, whereas certain high-probability leaves contribute less to the final decision. This discrepancy suggests that DSP performs implicit reweighting of decision evidence across multiple leaves, thereby enabling a more robust aggregation of complementary signals for final prediction.

\begin{figure}[!t]
\centering
\includegraphics[width=0.95\linewidth]{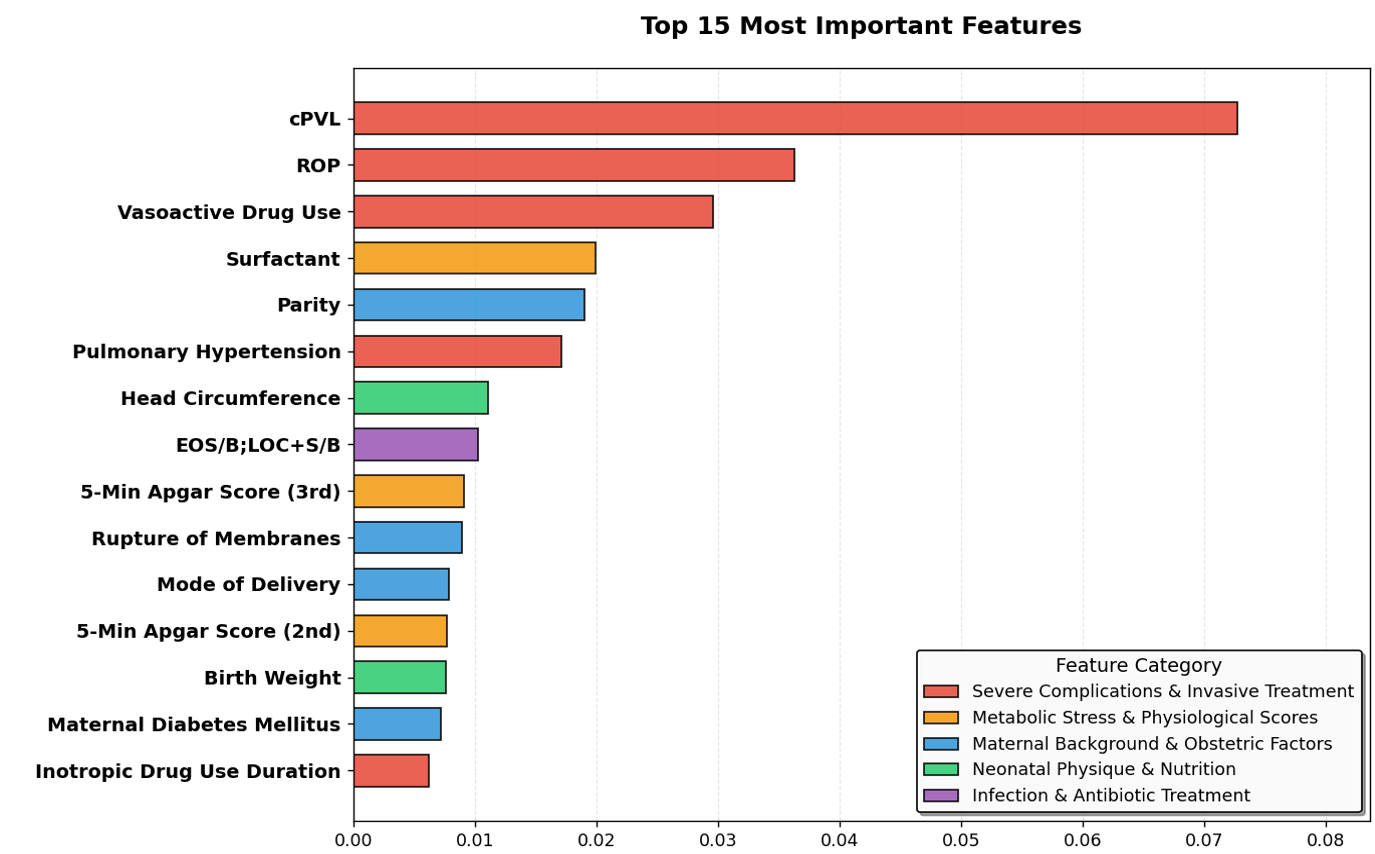}
\caption{Global feature importance derived from SHAP analysis on the test set.}
\label{fig8}
\end{figure}

We further examine global feature dependencies using SHAP analysis. Figure~\ref{fig8} presents the global feature importance aggregated over the test set. The model exhibits a structured and clinically consistent feature hierarchy. Among all variables, cPVL emerges as the most influential predictor, indicating its dominant role in outcome determination. Severe neonatal complications, including retinopathy of prematurity (ROP), pulmonary hypertension, and vasoactive drug usage, also contribute substantially, reflecting their strong association with adverse neurodevelopmental outcomes. Therapeutic and physiological indicators, such as surfactant administration and the 5-minute Apgar score, show moderate contributions, capturing early postnatal physiological stability. In addition, anthropometric measures, including birth weight and head circumference, provide consistent baseline risk information. Maternal factors, such as parity, exhibit relatively smaller but non-negligible effects. Overall, the global SHAP analysis confirms that the model primarily relies on clinically meaningful indicators reflecting disease severity, treatment intensity, and baseline neonatal condition, aligning well with established neonatal intensive care knowledge.

Finally, to provide case-level interpretability, \textbf{Appendix Fig.~\ref{fig:app_local_shap}} presents local SHAP explanations for representative high-confidence true-negative and true-positive samples. In the true-negative case, protective factors such as absence of cPVL, favorable anthropometric measures, and reduced hypoxic exposure collectively lower predicted risk. In contrast, in the true-positive case, pathological indicators such as pulmonary hypertension and infection-related complications significantly increase the predicted risk, complemented by additional contributions from perinatal and cardiovascular variables. Together, these local explanations are consistent with the global feature importance patterns, further demonstrating that QDSP captures clinically plausible and interpretable risk structures at both population and individual levels.

\section{Discussion}

The primary VLBWI cohort consists of 51 cases with 53 clinical variables, forming a high-dimensional and small-sample learning scenario characterized by substantial feature collinearity and inherent clinical heterogeneity. In such a regime, model robustness and stability are as critical as raw representational capacity. The proposed QDSP framework achieves strong and consistent performance under these constraints, reaching an AUC of 0.9714, accuracy of 0.9200, F1-score of 0.9416, and a Brier score of 0.0658 on the primary cohort, indicating both high discriminative ability and well-calibrated probabilistic outputs.

The ablation study further validates the internal design of QDSP. The QSS module contributes primarily to improved feature stability by reducing redundancy and enhancing the reliability of feature subspaces under sampling variability. In contrast, the DSP module improves nonlinear interaction modeling through differentiable structured decision learning, enabling flexible yet interpretable partitioning of the feature space. Notably, removing leaf-probability fusion (LPF) can lead to improvements in certain calibration-related metrics, but it generally reduces classification balance, suggesting that LPF plays a critical role in integrating complementary decision evidence across leaves. The full configuration, which combines QSS, DSP, and LPF, achieves the most balanced overall performance, highlighting the importance of jointly modeling feature stability and structured decision interactions.

Across external benchmarking datasets and progressively reduced training settings, QDSP demonstrates consistent robustness under data scarcity. This stability can be attributed to the integration of stability-aware subspace sampling and structured probabilistic decision modeling, which together reduce sensitivity to sampling fluctuations and mitigate overfitting in small-sample regimes. Moreover, the model maintains strong interpretability through multiple complementary perspectives, including differentiable decision paths, leaf-level contribution decomposition, and SHAP-based global and local feature attribution, all of which align well with established clinical knowledge in neonatal medicine.

Despite these promising results, several limitations remain. First, the primary cohort is derived from a single center, which may limit population diversity and external validity. Although validation on public datasets partially mitigates this concern, true clinical generalization requires multicenter prospective studies. Second, the current framework relies exclusively on static variables collected at discharge, without incorporating longitudinal temporal dynamics that may further improve prognostic accuracy. Future work will focus on extending QDSP to multicenter datasets, integrating temporal and multimodal clinical data, and improving probability calibration and interpretability interfaces to facilitate real-world clinical deployment.

\section{Conclusion}

This study presents and validates the QDSP framework for early prediction of two mutually exclusive adverse outcomes, death and cerebral palsy, in very low birth weight infants (VLBWI). By integrating Quota-guided Subspace Sampling (QSS) with Differentiable-decision-guided Structure Perception (DSP), the proposed framework is specifically designed to address the challenges of high-dimensional, small-sample clinical tabular learning, where feature instability, nonlinear interactions, and limited data availability jointly hinder reliable prognostic modeling.

On the primary clinical cohort, QDSP achieves strong predictive performance, reaching an accuracy of 0.9200 and an AUC of 0.9714, and consistently outperforms representative state-of-the-art baselines including XGBoost, TabNet, and TabPFN. In addition, robustness and ablation studies demonstrate that the proposed architecture maintains stable discriminative performance under varying degrees of data scarcity, highlighting the effectiveness of jointly modeling feature stability (QSS) and structured probabilistic decision learning (DSP with LPF).

Beyond predictive performance, QDSP provides clinically meaningful interpretability through differentiable decision-path modeling, leaf-level contribution analysis, and SHAP-based global and local explanations. The identified key predictors, such as cPVL and birth weight, are highly consistent with established neonatal pathophysiological knowledge, supporting the clinical credibility of the model. Overall, QDSP offers a practical and interpretable framework for discharge-time risk stratification in VLBWI and may facilitate more precise and personalized intervention strategies for high-risk neonates.

\section*{Acknowledgments}
This work was supported by the National Natural Science Foundation of China under Grant 62006165 and the Natural Science Foundation of Sichuan Province under Grants 2025ZNSFSC1477 and 2024YFFK0077. Ling Wang and Xiaolong Li contributed equally to this work. Dapeng Chen and Nan Mu are corresponding authors.

\appendix
\setcounter{table}{0}
\setcounter{figure}{0}
\renewcommand{\thetable}{A\arabic{table}}
\renewcommand{\thefigure}{A\arabic{figure}}
\renewcommand{\theHtable}{appendix.table.\arabic{table}}
\renewcommand{\theHfigure}{appendix.figure.\arabic{figure}}
\section{Supplementary Experimental Results and Interpretability Analyses}

\begin{figure}[H]
\centering
\includegraphics[width=\textwidth]{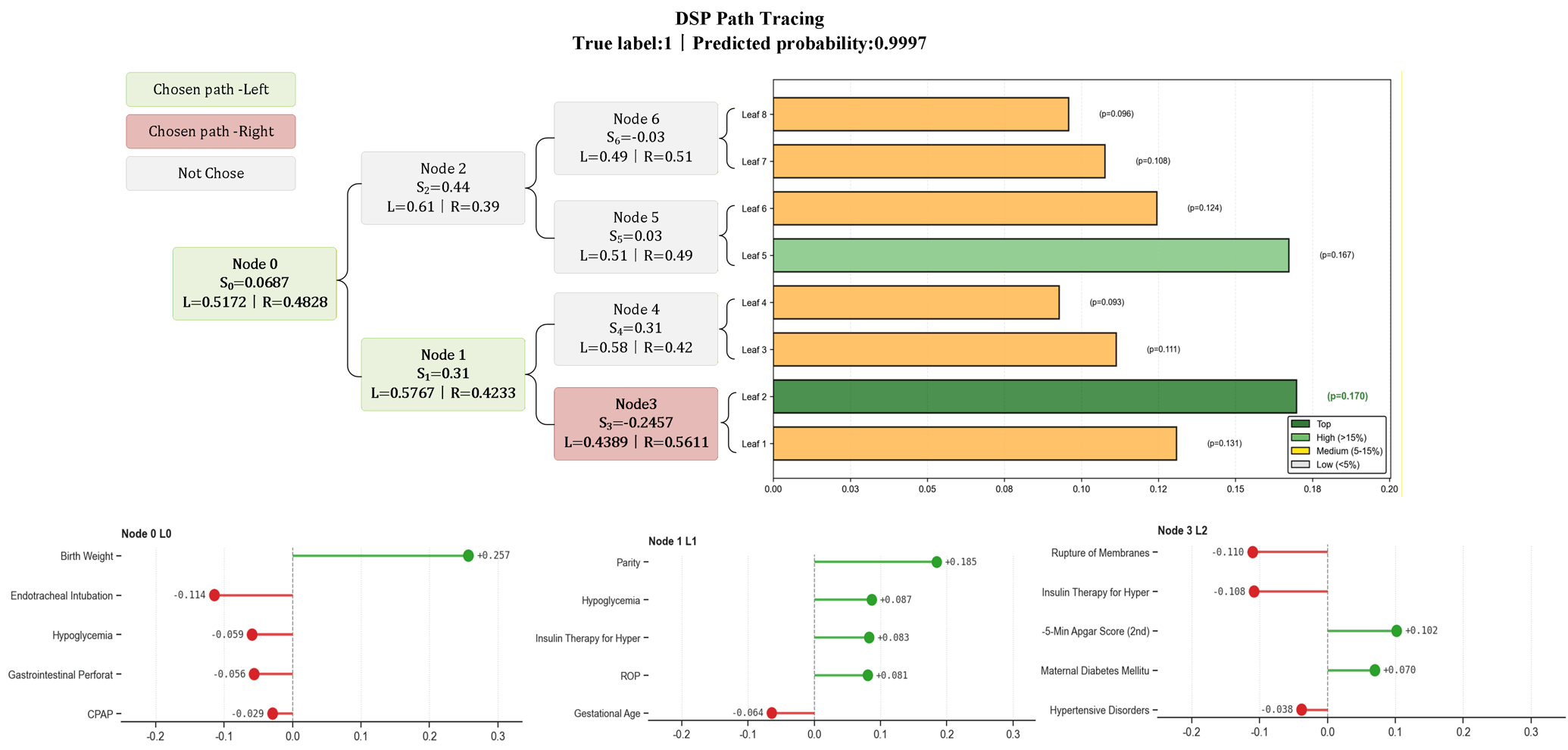}
\caption{DSP decision trace visualization.}
\label{fig:app_decision_trace}
\end{figure}

\begin{figure}[H]
\centering
\includegraphics[width=\textwidth]{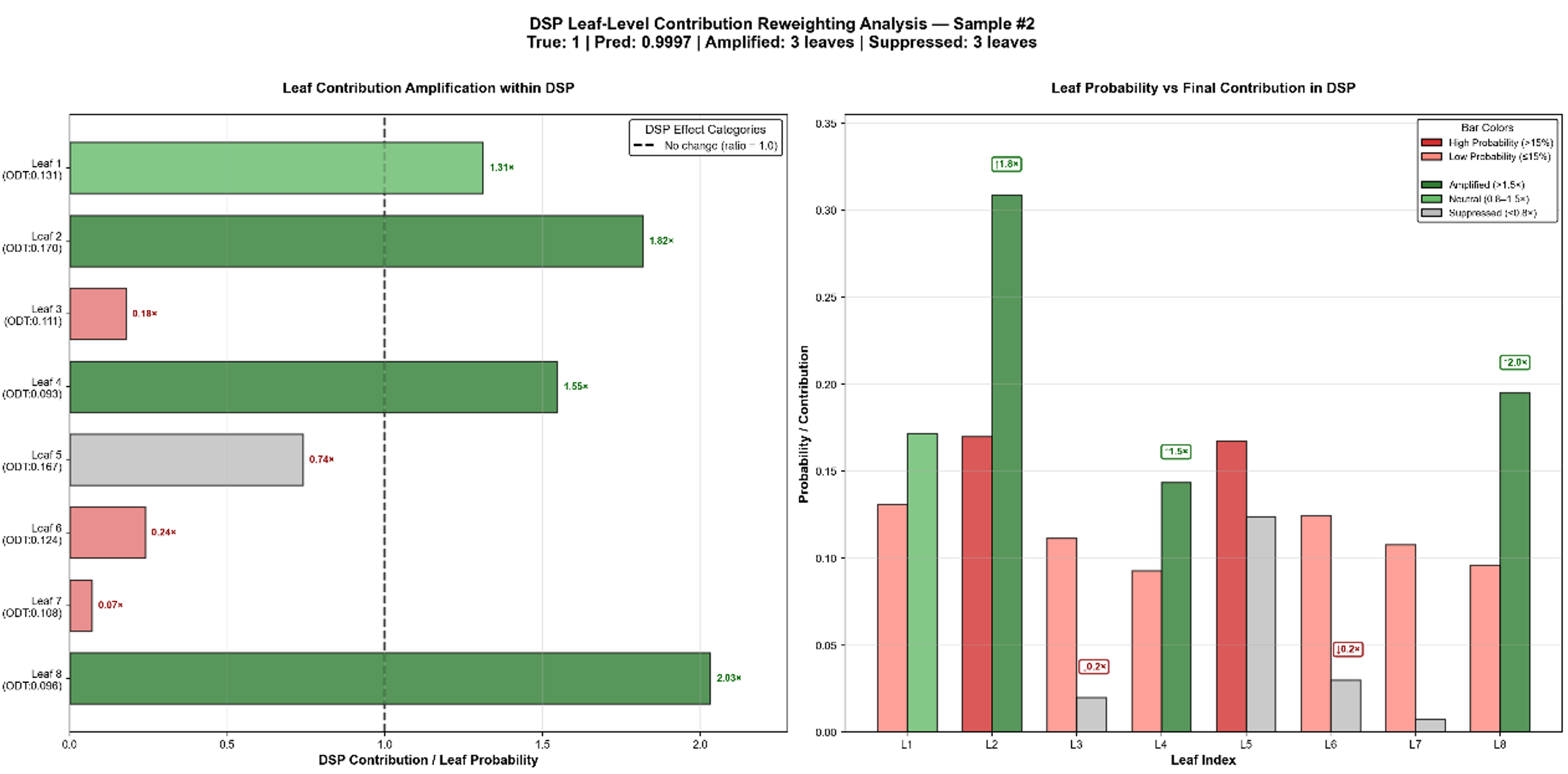}
\caption{DSP collaborative multi-leaf decision mechanism.}
\label{fig:app_multi_leaf}
\end{figure}

\begin{figure}[H]
\centering
\includegraphics[width=\textwidth]{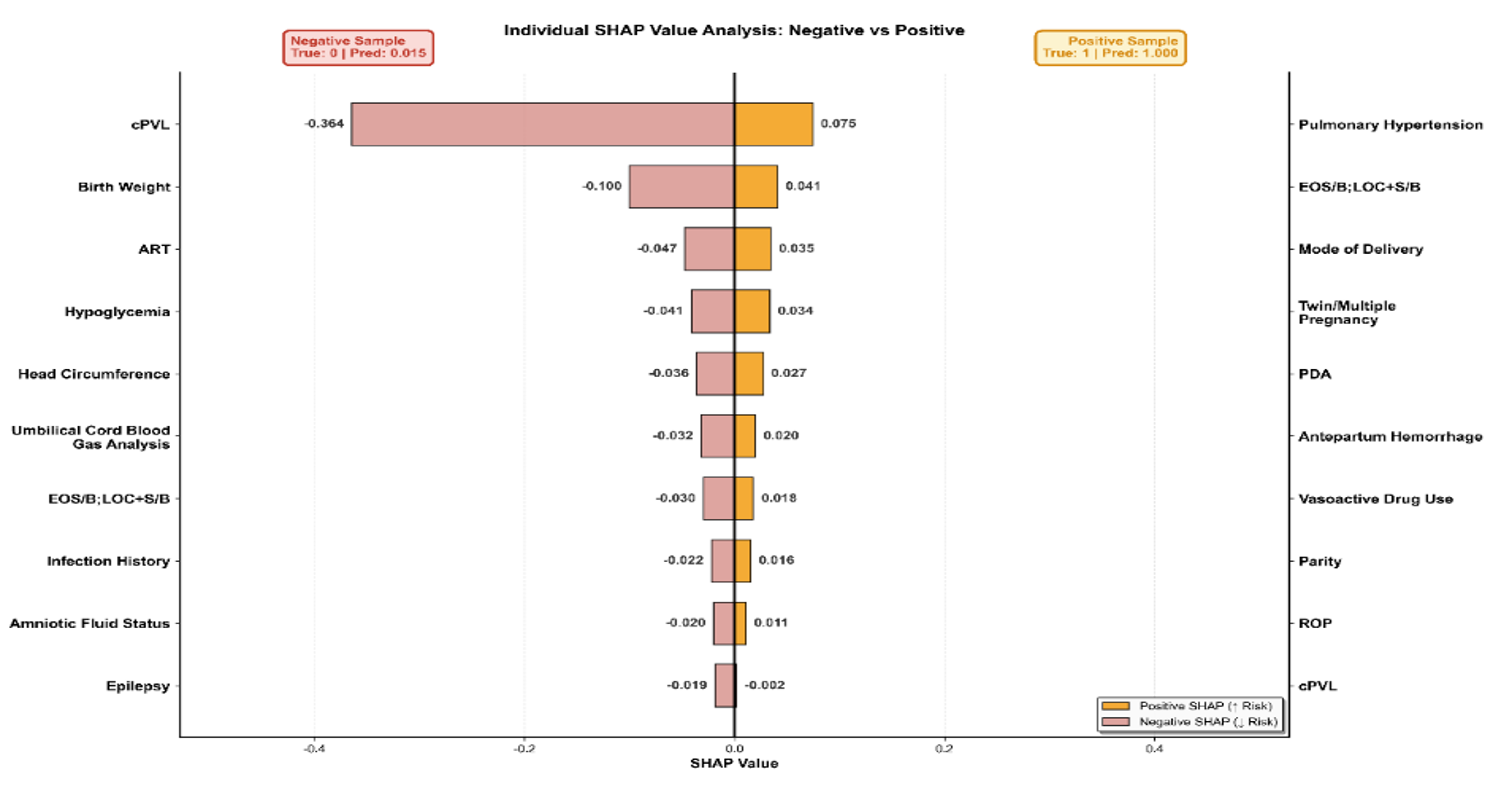}
\caption{SHAP-based feature attribution analysis for representative prediction cases.}
\label{fig:app_local_shap}
\end{figure}

This appendix provides additional experimental benchmarking results and interpretability analyses referenced in the main manuscript. Appendix Tables~\ref{tab:app_heart_performance}--\ref{tab:app_stroke_performance} present detailed benchmarking results of QDSP and competing baseline methods on three external public medical tabular datasets, including Heart Disease, Pima Diabetes, and Stroke Prediction. Appendix Figs.~\ref{fig:app_decision_trace}--\ref{fig:app_local_shap} provide additional visualization analyses for the proposed DSP interpretability framework, including probabilistic routing traces, collaborative multi-leaf decision behaviors, and SHAP-based patient-level feature attribution patterns.

\begin{table}[H]
\centering
\caption{Performance comparison on the Heart Disease dataset.}
\label{tab:app_heart_performance}
\small
\setlength{\tabcolsep}{4pt}
\renewcommand{\arraystretch}{1.1}
\resizebox{\textwidth}{!}{%
\begin{tabular}{lcccccc}
\toprule
Model & Accuracy$\uparrow$ & Precision$\uparrow$ & Recall$\uparrow$ & AUC$\uparrow$ & F1-Score$\uparrow$ & Brier Score$\downarrow$ \\
\midrule
Logistic Regression & 0.8265 & 0.7311 & \textbf{0.8208} & 0.8864 & \underline{0.7733} & 0.1221 \\
SVM                 & 0.8232 & 0.7507 & 0.7825 & \textbf{0.9064} & 0.7609 & 0.1184 \\
Decision Tree       & \underline{0.8367} & 0.7736 & 0.7736 & 0.8872 & \textbf{0.7736} & 0.1212 \\
Random Forest       & 0.8255 & 0.7444 & 0.7882 & 0.8923 & 0.7657 & 0.1287 \\
LightGBM            & 0.8197 & 0.7523 & 0.7636 & 0.8757 & 0.7517 & 0.1278 \\
XGBoost             & 0.8070 & 0.7149 & \underline{0.8056} & 0.8832 & 0.7502 & 0.1243 \\
TabNet              & 0.8164 & 0.7784 & 0.7275 & \underline{0.9039} & 0.7401 & \underline{0.1145} \\
TabPFN              & 0.8162 & \underline{0.7873} & 0.6684 & 0.8939 & 0.7212 & 0.1246 \\
QDSP (Ours)         & \textbf{0.8435} & \textbf{0.8081} & 0.7446 & 0.9036 & \textbf{0.7736} & \textbf{0.1086} \\
\bottomrule
\end{tabular}}
\end{table}

\begin{table}[H]
\centering
\caption{Performance comparison on the Pima Diabetes dataset.}
\label{tab:app_pima_performance}
\small
\setlength{\tabcolsep}{4pt}
\renewcommand{\arraystretch}{1.1}
\resizebox{\textwidth}{!}{%
\begin{tabular}{lcccccc}
\toprule
Model & Accuracy$\uparrow$ & Precision$\uparrow$ & Recall$\uparrow$ & AUC$\uparrow$ & F1-Score$\uparrow$ & Brier Score$\downarrow$ \\
\midrule
Logistic Regression & 0.6953 & 0.5644 & 0.7951 & 0.8149 & 0.6495 & 0.1749 \\
SVM                 & 0.7272 & 0.5858 & \underline{0.8096} & 0.8262 & 0.6747 & \underline{0.1614} \\
Decision Tree       & 0.6953 & 0.5644 & 0.7951 & 0.8149 & 0.6495 & 0.1749 \\
Random Forest       & 0.7350 & 0.5954 & 0.8060 & \underline{0.8351} & \textbf{0.6825} & 0.1642 \\
LightGBM            & 0.7389 & 0.5988 & 0.7967 & 0.8270 & 0.6816 & 0.1656 \\
XGBoost             & 0.7220 & 0.5752 & 0.8022 & 0.8214 & 0.6688 & 0.1706 \\
TabNet              & 0.7252 & 0.5880 & \textbf{0.8266} & \textbf{0.8358} & 0.6815 & \textbf{0.1575} \\
TabPFN              & \underline{0.7532} & \underline{0.6140} & 0.6863 & 0.8061 & 0.6481 & 0.1641 \\
QDSP (Ours)         & \textbf{0.7573} & \textbf{0.6285} & 0.7430 & 0.8212 & 0.6809 & 0.1908 \\
\bottomrule
\end{tabular}}
\end{table}

\begin{table}[H]
\centering
\caption{Performance comparison on the Stroke Prediction dataset.}
\label{tab:app_stroke_performance}
\small
\setlength{\tabcolsep}{4pt}
\renewcommand{\arraystretch}{1.1}
\resizebox{\textwidth}{!}{%
\begin{tabular}{lcccccc}
\toprule
Model & Accuracy$\uparrow$ & Precision$\uparrow$ & Recall$\uparrow$ & AUC$\uparrow$ & F1-Score$\uparrow$ & Brier Score$\downarrow$ \\
\midrule
Logistic Regression & \underline{0.7457} & \textbf{0.6821} & 0.7671 & 0.8331 & 0.7221 & \underline{0.1667} \\
SVM                 & 0.7370 & 0.6655 & 0.7831 & 0.8089 & 0.7196 & 0.1772 \\
Decision Tree       & 0.6816 & 0.6127 & 0.7186 & 0.7508 & 0.6588 & 0.2096 \\
Random Forest       & 0.7249 & 0.6642 & 0.7309 & 0.8066 & 0.6960 & 0.1816 \\
LightGBM            & 0.7336 & 0.6498 & 0.8273 & 0.8319 & 0.7279 & 0.2060 \\
XGBoost             & 0.6263 & 0.5383 & \textbf{0.9317} & 0.7996 & 0.6824 & 0.2187 \\
TabNet              & 0.7370 & 0.6474 & \underline{0.8554} & \textbf{0.8455} & 0.7370 & 0.1698 \\
TabPFN              & 0.7328 & 0.6338 & 0.9000 & 0.8279 & \textbf{0.7438} & 0.1690 \\
QDSP (Ours)         & \textbf{0.7526} & \underline{0.6721} & 0.8331 & \underline{0.8371} & \underline{0.7433} & \textbf{0.1666} \\
\bottomrule
\end{tabular}}
\end{table}

\end{document}